\documentclass{article}
\usepackage{spconf, amsmath, graphicx, media9, animate}
\usepackage{booktabs}
\usepackage{multirow}
\usepackage{hyperref}
\usepackage{cite}

\title{Piano Skills Assessment}
\name{Paritosh Parmar$^{1 *}$ \hspace{1cm} Jaiden Reddy$^{2 *}$ \thanks{$^{*}$Equal contribution.} \hspace{1cm} Brendan Morris$^{1}$}
\address{$^{1}$University of Nevada, Las Vegas \hspace{1cm} $^{2}$Ed W. Clark High School, Las Vegas}
\begin{document}
%
\maketitle
\begin{abstract}
Can a computer determine a piano player's skill level? Is it preferable to base this assessment on visual analysis of the player's performance or should we trust our ears over our eyes? Since current CNNs have difficulty processing long video videos, how can shorter clips be sampled to best reflect the players skill level? In this work, we collect and release a first-of-its-kind dataset for multimodal skill assessment focusing on assessing piano player's skill level, answer the asked questions, initiate work in automated evaluation of piano playing skills and provide baselines for future work.
\end{abstract}
%
\section{Introduction}
\label{sec:introduction}
Automated evaluation of skills/action quality involves quantifying \textit{how skillful the person is at the task at hand}/ or \textit{how well the action was performed}. Automated skills assessment (SA)/action quality assessment (AQA) is needed in a variety of areas including sports judging and even education as has been recently underscored due to the ongoing COVID-19 pandemic which has severely reduced in-person teaching and guidance. Moreover, automated evaluation of skills can make learning more accessible to socioeconomically disadvantaged subgroups of our society. Apart from teaching and guidance, it can be employed in video retrieval, and can provide second opinions in the case of controversial judging decisions in order to remove biases.

In this paper, we address automated determination of piano playing skill level based on a 10 point scale using a new multimodal PIano Skills Assessment (PISA) dataset (Fig. \ref{fig:dataset_samples}) that accounts for both visual and auditory evidence.
\section{Related Work}

\textbf{Piano + Computer Vision:} There has been work exploring the use of computer vision in determining the accuracy of pianists \cite{lee2019observing, takegawa2006design, gorodnichy2006detection, oka2013marker, suteparuk2014detection}; automatic transcription \cite{akbari2015real, deb2016image, li2018creating, koepke2020sight, li2020robust}; generating audio from video/silent performance \cite{kang2019virtual, su2020audeo, gan2020foley}; or generating pianist's body pose from audio/MIDI \cite{li2018skeleton}. However, there has been no work addressing the prediction of pianist's skill level from the recording of their performance. Our work is the first to address prediction of pianist's skill level in an automated fashion. 

\noindent\textbf{Skills-Assessment/Action Quality Assessment:} While there has been quite some progress recently in the areas of SA \cite{doughty2018s, li2018evaluation, doughty2019pros, li2019manipulation, seo2019understanding, li2019automated, wang2020towards, chen2020toward} and AQA \cite{gordon1995automated, pirsiavash2014assessing, parmar2016measuring, parmar2017learning, li2018end, xiang2018s3d, li2018scoringnet, xu2019learning, parmar2019action, lei2020learning, parmar2019hallucinet, sardari2019view, pan2019action, mtlaqa, jain2019unsupervised, ogata2019temporal, gao2020asymmetric, wang2020hands, wang2020assessing, zeng2020hybrid, nekoui2020falcons, tang2020uncertainty, du2020multi, jain2020action, sardari2020vi, condorivirtual, parsa2020multi}, none of these works address assessment of pianist's skills. Moreover, ours is the first work to take a multimodal learning approach in skills-assessment or action quality assessment.
\section{Multimodal PISA Dataset}
\label{sec:dataset}
\begin{figure}[]
\small
\centering
\setlength\tabcolsep{1pt}%
\begin{tabular}{lcccc}
& \includegraphics[width=0.22\columnwidth]{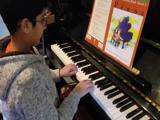}
& \includegraphics[width=0.22\columnwidth]{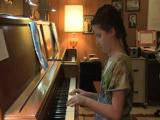}
& \includegraphics[width=0.22\columnwidth]{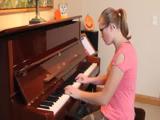}
& \includegraphics[width=0.22\columnwidth]{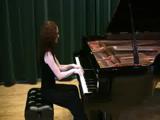}\\
& \animategraphics[loop,autoplay,poster=1,width=0.22\columnwidth]{30}{Figs/videos/hands/8/}{0000}{0399} 
& \animategraphics[loop,autoplay,poster=1,width=0.22\columnwidth]{30}{Figs/videos/hands/9/}{0000}{0399} 
& \animategraphics[loop,autoplay,poster=1,width=0.22\columnwidth]{30}{Figs/videos/hands/19/}{0000}{0399} 
& \animategraphics[loop,autoplay,poster=1,width=0.22\columnwidth]{30}{Figs/videos/hands/61/}{0000}{0599}  \\
& \includemedia[
  addresource="Figs/audios/8.mp3",
  transparent,
  flashvars={
    source="Figs/audios/8.mp3"
   &autoPlay=true
  },
]{\includegraphics[height=5ex]{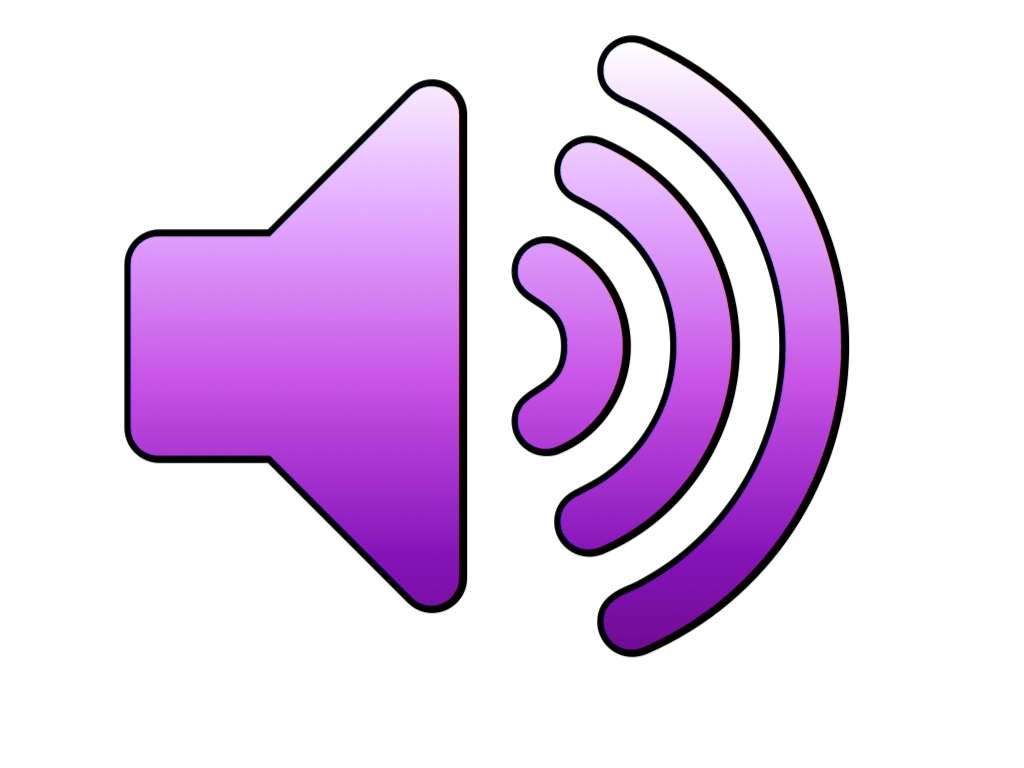}}{APlayer.swf} 
& \includemedia[
  addresource="Figs/audios/9.mp3",
  transparent,
  flashvars={
    source="Figs/audios/9.mp3"
   &autoPlay=true
  },
]{\includegraphics[height=5ex]{Figs/speaker_1.png}}{APlayer.swf} 
& \includemedia[
  addresource="Figs/audios/19.mp3",
  transparent,
  flashvars={
    source="Figs/audios/19.mp3"
   &autoPlay=true
  },
]{\includegraphics[height=5ex]{Figs/speaker_1.png}}{APlayer.swf} 
& \includemedia[
  addresource="Figs/audios/61.mp3",
  transparent,
  flashvars={
    source="Figs/audios/61.mp3"
   &autoPlay=true
  },
]{\includegraphics[height=5ex]{Figs/speaker_1.png}}{APlayer.swf} \\
\textbf{\textsc{SL}} & 2 & 2 & 6 & 9 \\
\textbf{\textsc{PL}} & 2 & 3 & 6 & 10
\end{tabular}
\caption{\textbf{Examples of samples from our dataset.} First row: images of full scenes; Second row: close-up videos of pianists' hands (only a part of the full sample shown); Third row: audio of the performances; Fourth row: song-level (\textsc{SL}) (1 to 10); Fifth row: player-level (\textsc{PL}) (1 to 10). \textit{To watch videos and listen to audios, please view this paper in Adobe Reader with multimedia settings enabled. Control audios with Right-Click}. The goal of this paper is to predict the player-level from video and/or audio.}
\label{fig:dataset_samples}
\end{figure}

The new PISA dataset includes two attributes in need of definition: \textbf{player skill} and \textbf{music difficulty}. A 10 point grading system was selected for player skill/level (PL) based on a technical and repertoire syllabus \cite{jaiden_1} developed by the local chapter of the Music Teachers National Association (MTNA). For certification services with other metrics, the grading system used in this paper can be roughly translated as follows: levels 1-9 represent pre-collegiate certification/skill level, while level 10 represents a collegiate or post-collegiate mastery. The primary evaluation of player skill is through the technique required for the most difficult song they are able to play. In this way, song level generally indicates player skill.

The methodology for determining the song level (SL) of difficulty of a piano piece utilized multiple syllabi. A 10 point grading system was used for piece difficulty. Songs rated in the Con Brio syllabus \cite{jaiden_3} as levels 1-8 were given a 1-8 rating, songs rated as pre-collegiate were given a 9 rating, and songs rated as collegiate or beyond were given a 10 rating. Songs not present in the Con Brio syllabus were referenced in the Henle syllabus \cite{jaiden_4}, which rates pieces from 1-9. Pieces at level 9 in the Henle syllabus were cross checked with the Royal Conservatory of Music syllabus \cite{jaiden_2} to determine if they would be assigned a 9 or a 10 in our dataset (because the distinction between 9 and 10 is not made in the Henle syllabus). 

\begin{figure}
    \centering
    \includegraphics[width=0.45\columnwidth]{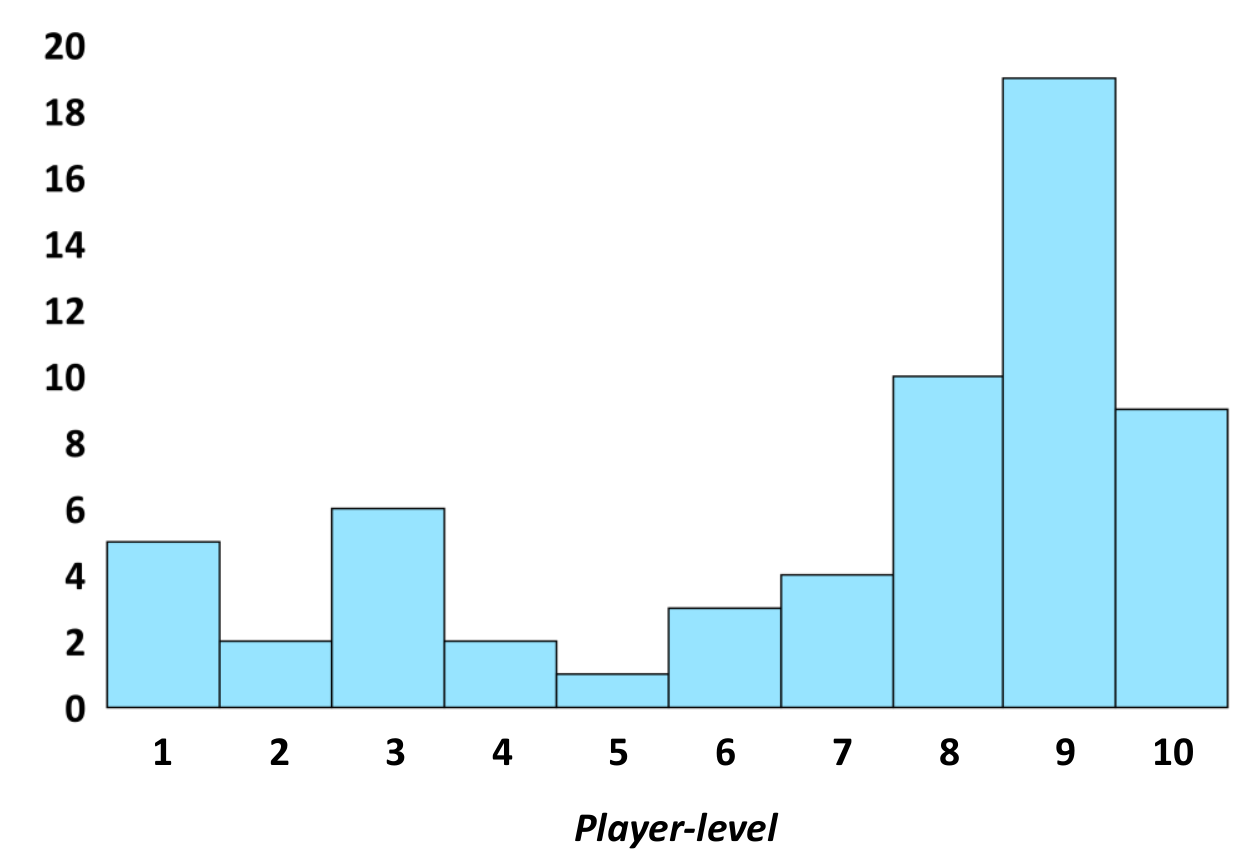}
    \includegraphics[width=0.45\columnwidth]{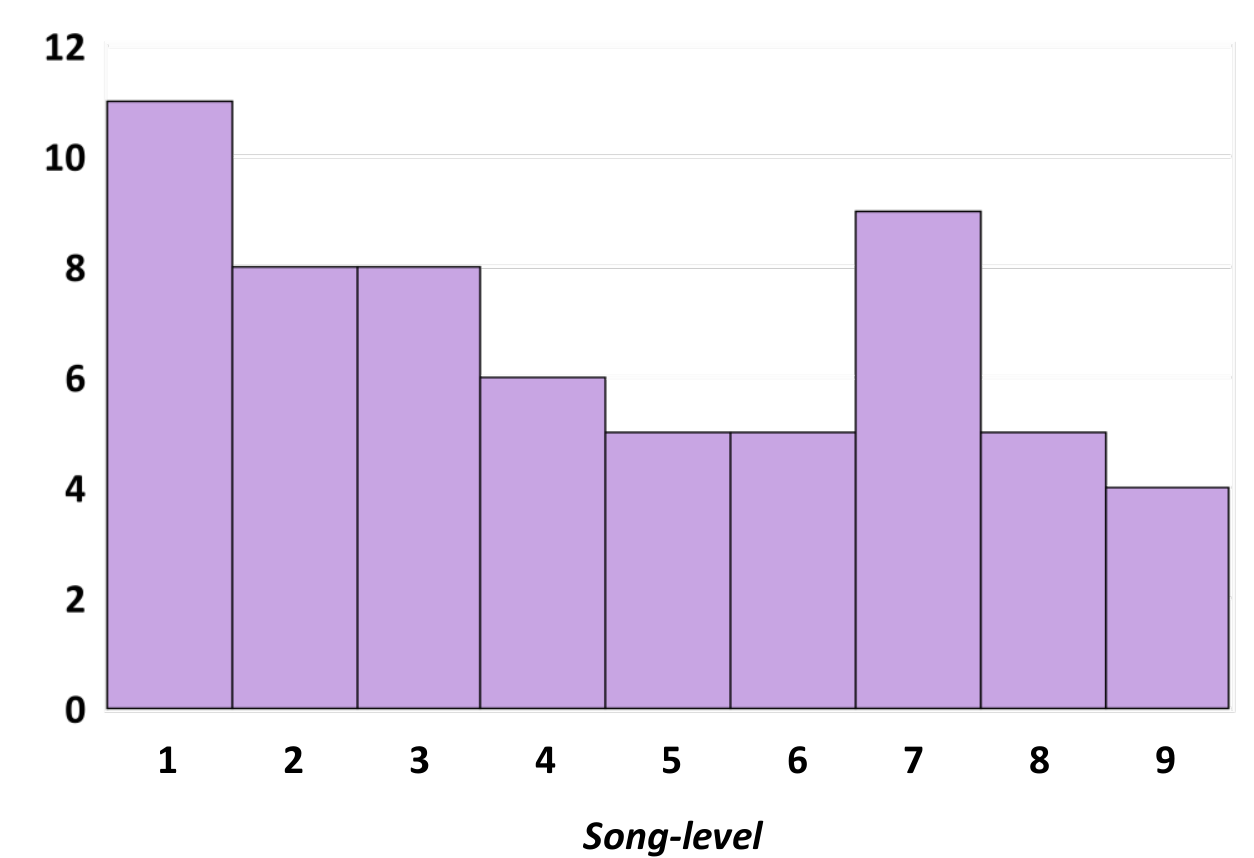}
    \caption{\textbf{Histograms of player-levels and song-levels.}}
    \label{fig:histograms}
\end{figure}

\begin{figure}
    \centering
    \includegraphics[width=0.9\columnwidth]{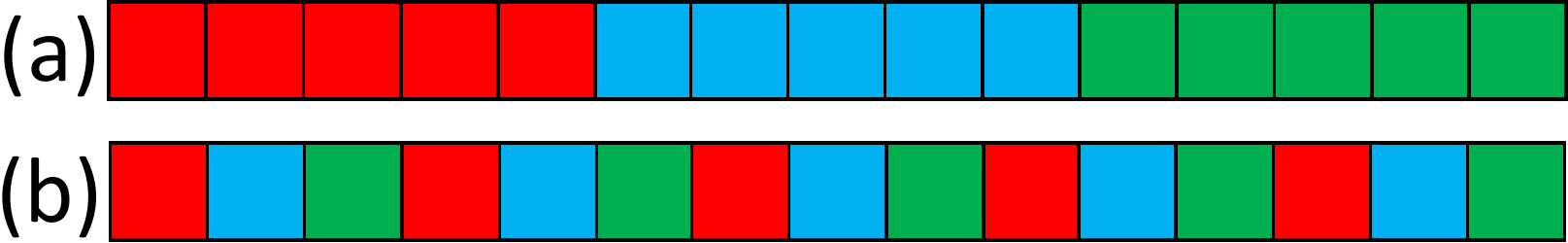}
    \caption{\textbf{Sampling schemes}: (a) Contiguous; (b) Uniformly Distributed. Time along the x-axis. Each color represents a sample. Each square represents a clip of 16 frames.}
    \label{fig:sampling_schemes}
\end{figure}

Existing skills assessment and action quality assessment have been prepared either through crowd-sourcing or direct annotation present in the video footage. Compared to those methods, we found collecting piano skills assessment data to be challenging. We had to rely on a trained pianist to collect videos from YouTube, analyze them, and determine each player’s skill level (as described above). As such, scaling up the dataset is difficult. We collected a total of 61 total piano performances. \textbf{Minimum, average, and maximum performance lengths} were \textbf{570, 2690, and 10038 frames}, respectively. Histograms with the no. of samples for each player-level and song-level are in Fig. \ref{fig:histograms}. To mitigate small dataset size, we create multiple \textit{unique}, \textit{non-overlapping} samples, each of size of \textbf{160 frames}. In this  way, we have a total of \textbf{992 unique samples}. We considered two types of sampling schemes: 1) \textbf{contiguous}; 2) \textbf{uniformly distributed}, which are illustrated in Fig. \ref{fig:sampling_schemes}. Out of the \textbf{61 total piano performances}, we use \textbf{31} (\textbf{516 samples}) as our \textbf{training set} and \textbf{30} (\textbf{476 samples}) as our \textbf{test set}. Note that no overlap exists between train and test sets. For each piano performance, we provide the following annotations: 1) \textbf{player skill level}; 2) \textbf{song difficulty level}; 3) \textbf{name of the song}; 4) \textbf{a bounding box around the pianist’s hands} (refer Fig. \ref{fig:dataset_samples}).

\section{Approach}
\label{sec:approach}
\begin{figure*}
\centering
\includegraphics[width=0.95\textwidth]{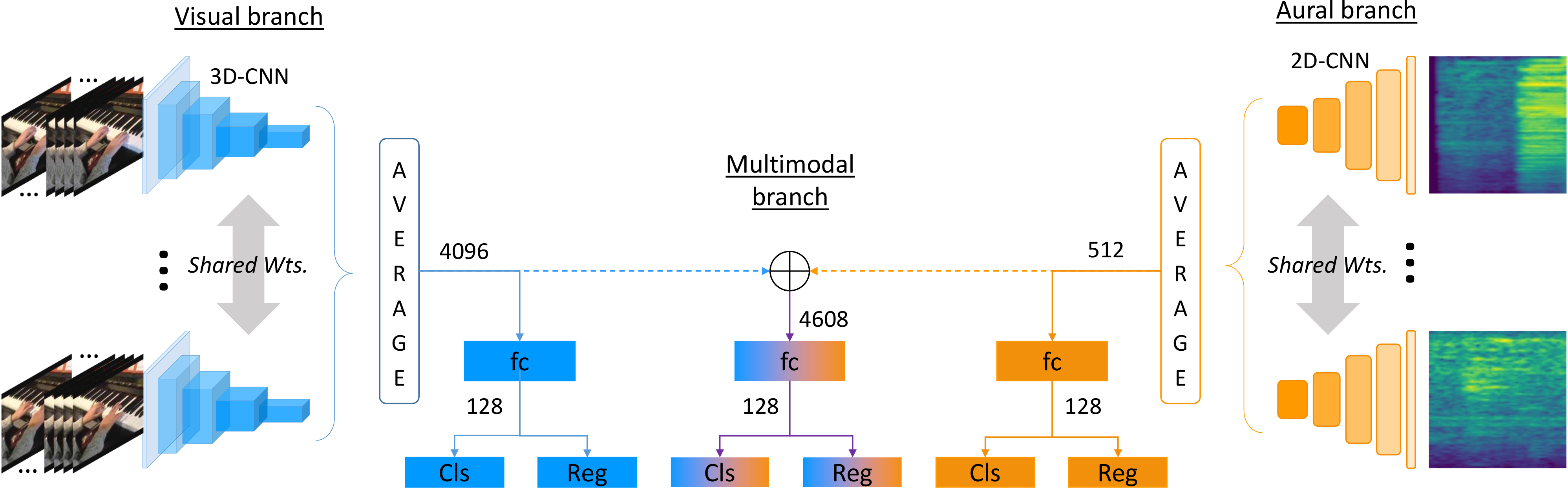} 
\caption{\textbf{Our multimodal learning architecture.} $\oplus$ represents concatenation operation.}
\label{fig:mmdl_arch}
\end{figure*}
In this section, we first present unimodal SA approaches, which are followed by a multimodal approach. Our full framework is shown in Fig. \ref{fig:mmdl_arch}. Finally, we provide our learning objective.

\subsection{Visual branch}
While initially, judging music visually rather than auditorily seems impossible, there are skills that can be judged visually. These skills would generally fall into two categories - technical skills, which build from grade to grade (in the above syllabi), and virtuoistic skills or professional skills, which are skills unique to a pianist that would be present in only grade 9 or 10 pianists. The first category includes the difficulty and speed of the scales and arpeggios included in the music. For example, the level 10 syllabus of the Las Vegas Music Teachers Association Technical Portion \cite{jaiden_5} includes all major scales (four octaves) and all minor scales (four octaves, natural,  harmonic, and melodic included). In a visual examination of a grade 10 song, such as Chopin’s Etude Op 25 No 11, the four octave A harmonic minor scale at high speed would immediately indicate a grade 10 piece. The same logic can be applied to arpeggios and cadences. This would cover the first category; the second category are skills unique to a pianist. The lack of such skills would not give any indication as to the level of a pianist, but the presence of such skills would immediately indicate a very high level of technical achievement. For example, professional pianists who must play at high speeds may play eight note intervals with their first and third fingers, which is incredibly difficult for the average pianist.

In order to take into account the factors mentioned, we believe processing short clips with 3DCNNs would be more suitable than processing single frames using 2DCNNs. Since we sample several such clips and process them individually using 3DCNNs, we need to aggregate individual clip-level features. Prior work \cite{mtlaqa} in AQA has shown averaging to work well and it also enables end-to-end training. RNN based aggregation would not enable end-to-end training when used with 3DCNNs due to large number of parameters and consequent overfitting. Therefore, we choose to use averaging as our aggregation scheme to obtain whole sample-level video features from clip-level features.

When considering just the visual branch, we deactivate everything other than the blue colored portion Fig. \ref{fig:mmdl_arch}. We further pass the whole sample-level video features through a linear layer to reduce its dimensions to 128.

\subsection{Aural branch}
A significant amount of information can be detected from audio as well. The velocity of the music (notes per second) can be detected auditorily, which is a simple yet valuable tool to judge the technical skill required for a piece. The presence of multiple notes at once with cadences that correspond to cadences included in the grade level syllabi above would be recognizable both through auditory and visual analysis. Unfortunately, the significant variety in style, clarity, and dynamics from song to song, while an imperative judge of piano skill in competitions and performances by judges familiar with said songs, makes an unfamiliar judge or computer system less effective at determining skill.

We convert the raw audio signal corresponding to sampled clips (the same clips as those mentioned in the visual branch) to its melspectrogram, which we then process using a 2DCNN. Similar to the visual branch, we aggregate the clip-level audio features using an averaging function. We learn the parameters of the 2DCNN end-to-end. 

When considering just the aural branch, we deactivate everything other than orange colored part in Fig. \ref{fig:mmdl_arch}. We further pass the whole sample-level audio features through a linear layer to reduce its dimensions to 128.

\subsection{Multimodal branch}
In the multimodal branch, we take the video-level video and audio features and concatenate them to produce multimodal features. We further pass the whole sample-level feature through a linear layer to reduce its dimensions to 128. We do not back-propagate from the multimodal branch to individual modality backbones to avoid cross-modality contamination.

\subsection{Objective function}
Unlike a typical classification problem, in our player-level prediction problem, the distance between categories has meaning. For example, for a ground-truth player-level of 5, although predicted levels of 2 and 6 are both incorrect, a predicted level of 6 is ``less wrong" than predicted level of 2. To address this characteristic, we incorporate a distance function ($\mathcal{L}_{Reg}$) in the objective in addition to cross-entropy loss ($\mathcal{L}_{CE}$). Specifically, we use a sum of L1 and L2 distances as our $\mathcal{L}_{Reg}$. L1 was found to be beneficial in \cite{mtlaqa}. Overall objective function is as shown in Eq. \ref{eq:overall_objective} with superscript of $V$ for visual and $A$ for audio cues. 

\begin{multline}
\label{eq:overall_objective}
    \mathcal{L}_{total} = \alpha_{1}\mathcal{L}_{CE}^{V} + \alpha_{2}\mathcal{L}_{Reg}^{V}\\ + \beta_{1}\mathcal{L}_{CE}^{A} + \beta_{2}\mathcal{L}_{Reg}^{A} + \gamma_{1}\mathcal{L}_{CE}^{M} + \gamma_{2}\mathcal{L}_{Reg}^{M}
\end{multline}
\section{Experiments}
\textbf{Preprocessing:} Visual information pertaining to the whole scene might not be useful in determining the player level. Instead, we crop and use visual information pertaining to the forearms, hands, and the piano as shown in Fig. \ref{fig:dataset_samples}. Using librosa \cite{librosa}, we convert the audio signal to its melspectrogram (settings adopted from \cite{hasith}), and express that information in decibels. 

\noindent\textbf{Implementations details:} We use PyTorch \cite{pytorch}  to implement our networks. We use an ADAM optimizer \cite{adam} with a learning rate of 0.0001 and train all the networks for 100 epochs, with a batch size of 4. In Eq. \ref{eq:overall_objective}, we set the values of $\alpha_{1}, \beta_{1}, \text{and} \gamma_{1}$ to 1, while those of $\alpha_{2}, \beta_{2}, \text{and} \gamma_{2}$ to 0.1. Codebase would be made publicly available.

\noindent\textbf{Visual branch:} Since our dataset is small, we design a small, custom 3DCNN network (\texttt{C64, MaxP, C128, MaxP, C256, MaxP, C256, MaxP, C256, MaxP}) to process the visual information. To avoid overfitting, we pretrain this 3DCNN on UCF101 action recognition dataset \cite{ucf101}. We use 16 consecutive frames to form the input clip for our 3DCNN. All the frames are resized to 112 $\times$ 112 pixels. We apply horizontal flipping during training.

\noindent\textbf{Aural branch:} We use ResNet-18 (R18) \cite{resnet} as the backbone for our aural branch. We found that initializing weights by ImageNet \cite{imagenet} helped a lot. We change the number of input channels of the first convolutional layer of the pretrained network from 3 to 1. We convert the melspectrogram of audio signals to single channel images and resize these images to 224 $\times$ 224 pixels before use with R18. We found that applying random cropping hurt the performance. This may be because the useful information is present in the lowest and highest frequencies and removing those in the process of cropping adversely affects the performance. 

We conduct experiments to answer the following questions: 
\begin{enumerate}
    \item Is it possible to determine the pianist's skill level using machine learning/computer vision? 
    \item What is the better sampling strategy: contiguous or uniform distribution? 
    \item Is a multimodal assessment better than a unimodal assessment? 
\end{enumerate}
\noindent\textbf{Performance Metric:} We use accuracy (in \%) as the performance metric. \\
\begin{table}[]
\small
\centering
\begin{tabular}{@{}lcc@{}}
\toprule
\multirow{2}{*}{\textbf{Modality}} & \multicolumn{2}{c}{\textbf{Sampling Scheme}}   \\ \cmidrule(l){2-3} 
                                   & \textbf{Contiguous} & \textbf{Uniformly Dist.} \\ \cmidrule(r){1-1} \cmidrule(l){2-2} \cmidrule(l){3-3} 
Video                              & 65.55                 & \underline{73.95}                        \\
Audio                              & 53.36                 & \underline{64.50}                        \\
\textsc{MMDL}                               & 61.55                 & \textbf{\underline{74.60}}                        \\ \bottomrule
\end{tabular}
\caption{\textbf{Performance} (accuracy in \%) of single modalities vs a multimodal (\textsc{MMDL}) assessment for contiguous and uniformly distributed sampling schemes.}
\label{tab:res_smdl_vs_mmdl}
\end{table}

The results are presented in Table \ref{tab:res_smdl_vs_mmdl}. We observe that uniformly sampling is better than contiguous sampling. This may be due to the fact that different parts of the song can contain different elements, which can be better at providing a more diverse base for testing the pianist's skill set. Another finding to note is that visual analysis was better than aural analysis. Moreover, using both visual and aural features yielded the best results. As discussed in Sec. \ref{sec:approach}, different/unique elements of skills can be observed visually and aurally which may account for the small boost in performance over visual alone in the uniform sampling case. Results also show that we were able to train both the networks end-to-end on a small dataset, which justifies our design choices.

Furthermore, a significant gap in performances from different sampling schemes also suggests that the networks are not biased towards some trivial, low-level, local/static cues present in video or audio streams.
\section{Conclusion}
\label{sec:conclusion}
In this work, we addressed the automated assessment of piano skills. We introduced PISA the first ever piano skills assessment dataset. We found that assessing the performance on the basis of visual elements was better than on the basis of aural elements. However, our best performing model was the one that combined visual and aural elements. We found that uniformly distributed sampling was significantly better than contiguous sampling at reflecting a player's skill level. Computer vision is finding increasing applications in piano transcription, video to audio generation, etc. With our work, we hope to inspire future work in the direction of automated piano skills assessment and tutoring. For that, we have released our dataset and codebase and provided the performance baselines. While our approach yielded good results, we believe there is a significant scope for improvement in this direction.
%

\bibliographystyle{IEEEbib}
\bibliography{refs}

\end{document}